\documentclass[10pt, twocolumn, a4paper]{article}

\usepackage{amssymb}
\usepackage{graphicx}
\usepackage{amsmath}
\usepackage{amsfonts}
\usepackage{amssymb}
\usepackage{amsthm}
\usepackage{verbatim}
\usepackage{enumerate}
\usepackage{mdwlist} 
\usepackage{subfig}
\usepackage{color}
\usepackage[utf8]{inputenc}

\usepackage{algpseudocode}
\usepackage{algorithmicx}
\usepackage{lineno}
\usepackage{multirow}
\usepackage{array,url,kantlipsum}

\RequirePackage[left=2cm,%
                right=2cm,%
                top=2.25cm,%
                bottom=2.25cm,%
                headheight=12pt,%
                letterpaper]{geometry}%

\begin{document}

\title{Binary Distance Transform to Improve Feature Extraction}

\author{Mariane Barros Neiva$^{1,2}$ \footnote{marianeneiva@usp.br}, Antoine Manzanera$^3$ \footnote{antoine.manzanera@ensta-paristech.fr},
Odemir Martinez Bruno$^1$ \footnote{bruno@ifsc.usp.br}}

\date{\noindent{$^1$Scientific Computing Group, S\~ao Carlos Institute of Physics, University of S\~ao Paulo - USP, PO Box 369, 13560-970, São Carlos, S\~ao Paulo, Brazil - ttp://scg.ifsc.usp.br
\\
\noindent{$^2$Institute of Mathematics and Computer Science, University of S\~ao Paulo (USP), Avenida Trabalhador S\~ao Carlense,
400 13566-590 São Carlos, S\~ao Paulo, Brazil}
\\
\noindent{$^3$ENSTA-ParisTech, U2IS-Robotics \& Vision, \\Universit\'e de Paris-Saclay, 828, Boulevard des Mar\'echaux, 91762 Palaiseau CEDEX, France}}}

\maketitle

\begin{abstract}
To recognize textures many methods have been developed along the years. However, texture datasets may be hard to be classified due to artefacts such as a variety of scale, illumination and noise. This paper proposes the application of binary distance transform on the original dataset to add information to texture representation and consequently improve recognition. Texture images, usually in grayscale, suffers a binarization prior to distance transform and one of the resulted images are combined with original texture to improve the amount of information. Four datasets are used to evaluate our approach. For Outex dataset, for instance, the proposal outperforms all rates, improvements of an up to 10\%, compared to traditional approach where descriptors are applied on the original dataset, showing the importance of this approach.    
\end{abstract}



\section{Introduction} 
\label{sec:intro}

Texture although its difficulty of definition by the scientific community \cite{tuceryan1993texture} have been proved as an important source of information for pattern recognition \cite{kim1999statistical, haralick1979statistical, garcia2014local, costa2012efficient, guo2010completed, guo2010rotation, ojala2002multiresolution, journaux2008texture, daugman1988complete,florindo2016local,da2015feature}. In a classical recognition approach, images are first converted to a numerical vector that represent them with reduced dimensionality (the feature extraction step). Then, and artificial intelligence methods are used to compare, separate vectors and label images. However, if those unidimensional representations are not adequate, classification rates are low. Nevertheless, these bad representations affects not only classification but also segmentation and synthesis that also rely on this information. 

Therefore, feature extraction is a crucial step to create robust classification and must be performed wisely. For this reason, different methods were developed along the years to describe images based on different approaches such as its structure, statistics, complexity and spectrum \cite{materka1998texture}. 

LBP \cite{ojala2002multiresolution}, for instance, is one of the examples from statistical category. It is a local descriptor that uses windowed information to compute a histogram of patterns. Given the popularity of this method, extensions of it have been proposed such as  LBPV \cite{guo2010rotation}. This later algorithm adds contrast information of the image to improve the classical approach. The method proposed in \cite{florindo2016local} uses the LBP along with fractal dimension analysis as a way to extract texture features. Also, another example of this category, the GLCM \cite{haralick1979statistical} extract features from the co-occurrence matrix calculated by analyzes of neighboring pixels in an image.

A different approach proposed in \cite{gonccalves2016texture} where images are modeled as complex networks and features are extracted based on random walks on the graph. In \cite{da2015improved}, the authors use cellular automatta to corrode the image and extract feature from the tesselation. Also, fractal dimension have been proved as an important way to analyze texture as shown in \cite{florindo2016texture, florindo2012fractal,florindo2012comparative, florindo2013texture}

However, if the data is not good enough, these methods may not be able to work properly. One of the reasons for bad representation is that usually feature extraction methods depend on good quality images where important feature are clear. Unfortunately, real dataset suffers by many types of influences that input some difficulty to recognition. Those problems occur due to presence of different scales in intra-class images, illumination changes, bad quality cameras, noise, among others. 

Thus, to reduce these artifacts, one can preprocess original images aiming to enhance certain properties that are relevant to characterize the scene. In the literature, there are many filters proposed to reduce noise and improve image quality such as unsharp masks, Gaussian filters, image normalization, among others \cite{gonzalez2008digital}.

Nowadays, still few researches pay attention on the importance of this approach showing a lack of number of proposals in this area \cite{machado2013partial, nanni2015improving, gadermayr2016making}. Therefore, in order to fill this gap and to analyze the complexity of the surface, this paper uses the distance transform to produce different images and combine these images with original ones to extract feature. Distance transform is commonly used for tasks such skeletonization, cluster analyzes, morphological operations \cite{parker2010algorithms}, segmentation \cite{vincent1991watersheds}, robotics \cite{chin2001vision}, among others. However, it will be used in this research as an intermediate step between original dataset and feature extraction. 

Euclidean distance was chosen due to its ability to expand the image by calculating the distance to all non-image pixels to the surface keeping only stronger characteristics of the image as distance grows. Also, It can be roughly compared to multiscale analysis given its intrinsic feature of creating a image with different resolutions of the original one. Therefore, the goal is to create a set of images that shows the complexity of each group of gray level and use them to improve representation. Experiments are evaluated in four texture datasets: Brodatz, Outex, Vistex and Usptex and shows an advantage in results of proposed technique compared with the use of six feature extraction methods applied on the original dataset.

This paper is organized as follows: Section \ref{sec:rm} explains the preprocessing method used to improve the six descriptors, also briefly described in the section. Then, Section \ref{sec:proposal} illustrates how the proposed method works, explaining the combination of images, feature extraction methods and texture classification. Finally, Sections \ref{sec:results} and \ref{sec:conclusions} the results and conclusions of the work are presented respectively. 

\section{Related Methods}\label{sec:rm}

The main method to be explained in this paper is the distance transform algorithm. Also, six feature extraction methods will be combined with these new images to check if does unions improve or not image representation.

\subsection{Euclidean Distance Transform (EDT)}

The definition of distance transform algorithm is very simple yet complicated. It finds the minimum distance of background pixels to a region of interest \cite{rosenfeld1968distance}. The distance can be chosen according to the application but most of tasks use the Euclidean metric and it is the one used in this proposal. Euclidean distance of two pixels p and q is defined as:

\begin{equation}
d = \sqrt{(x_p -x_q)^2 + (y_p - y_q)^2}
\end{equation}

Usually, for binary images, region of interest are white pixels (1) while background pixels are valued as 0, then distance pixel p is then computed as:

\begin{equation}
I_{dist} (p) = min(p,F)
\end{equation}
where function $min$ returns the minimum distance of pixel at position p to a set of foreground pixels F. 

Finally, the simple definition disappears when one think about the computational cost of calculating all distances to find the minimum one. For a simple image with size $n$, a brute force algorithm must calculate the distance of a pixel to all foreground pixels to find the closest one. This approach would add to much time for a preprocessing step but fortunately many authors have proposed alternatives to reduce computational cost. The method proposed in \cite{maurer2003linear}, used in this paper, works faster reducing the dimension constructing a partial Voronoi diagram, the algorithm is proved linear according to the number of $n$ pixels in the image.

\begin{figure}%
  \centering  \subfloat[Original Image]{{\includegraphics[width=4cm, height = 3.5cm]{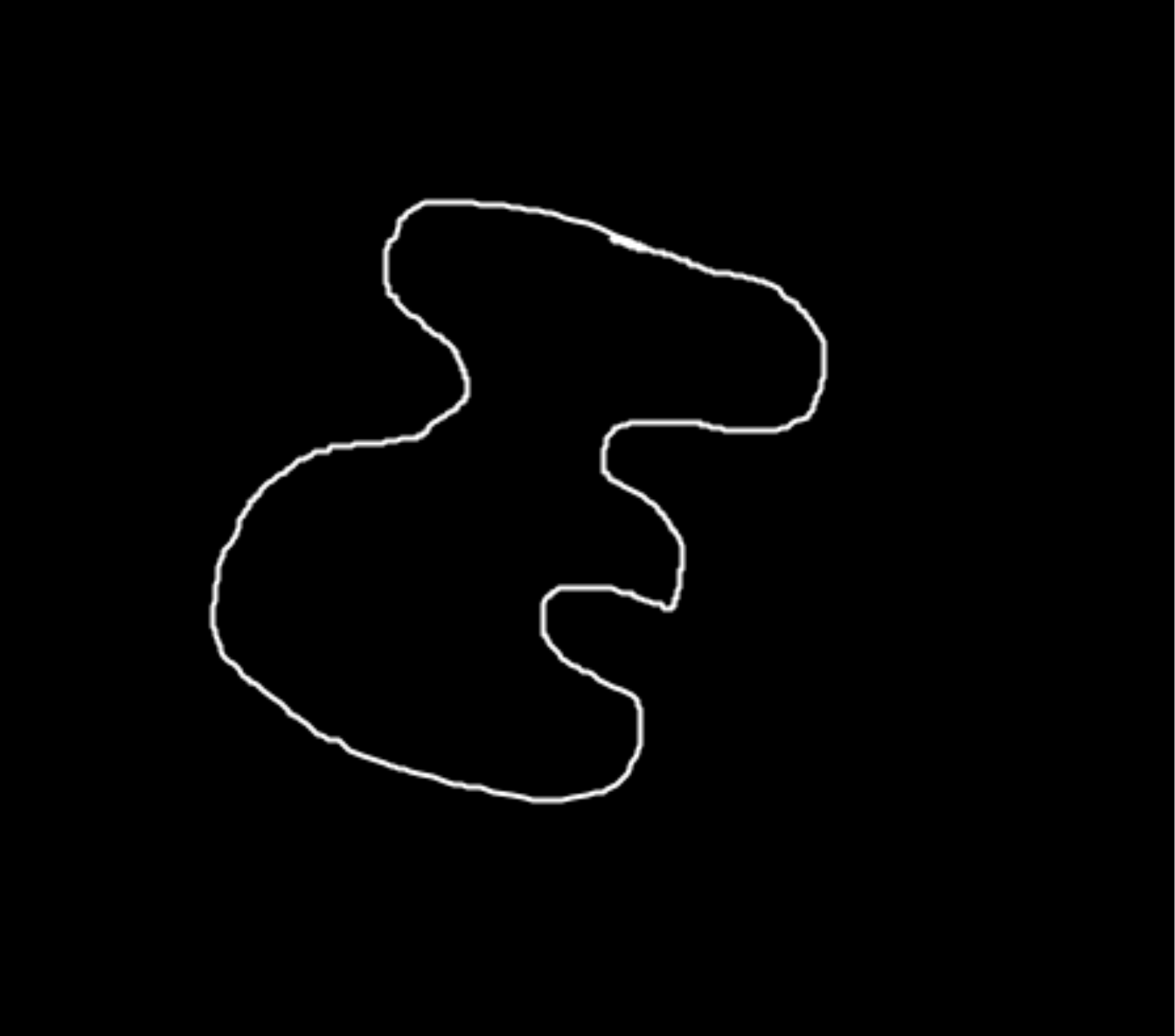} }}%
 \subfloat[Distance Image]{{\includegraphics[width=4cm,  height = 3.5cm]{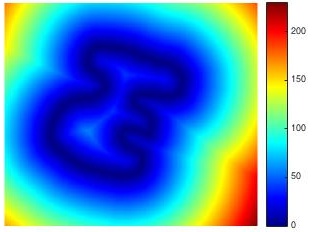} }}%
    \caption{Original image and respective distance transform (colors were added to improve visualization) } \label{fig:edt}
\end{figure}

Finally, to illustrate the transform, Figure \ref{fig:edt} shows an example of distance transform applied in a binary image. If we analyze pixels with the same distance in resulted image, it is possible to see that the approach dilates the initial surface and important features are maintained as long as distance grows. Therefore, the transformation enhances the complexity of objects in digitalized scene which can beneficiate feature extraction methods.

\subsection{Feature Extraction}\label{sec:features}

The goal of this proposal is to improve feature extraction. Therefore, six different methods were chosen to test the approach and check whether the addition of a transformed image improved or not the representation of images considering each descriptor. This section will briefly describe each of the methods used in the paper and its parameters.

\begin{itemize}
\item \textbf{Local Binary Pattern }(LBP): 

The simple but powerful descriptor analyses the image based on local patterns \cite{ojala2002multiresolution}. These patterns are obtained thresholding a window by its central pixel. When all pixels are used as central ones, MxN (size of the input image) pattern values are obtained and finally feature vector as the histogram of these patterns. The method requires two parameters to compute within a window: the radius $r$ and number of neighbors $P$. In this paper, we used the basic approach with r = 1 and P = 8. 

\item \textbf{Local Binary Pattern Variance} (LBPV): 

A variation of basic LBP is the rotation invariant local pattern descriptor represented by $LBP_{riu}$ \cite{ojala2000gray}. In this method, patterns that only contains two transitions from one to zero or zero to one and codes that only differ by rotation of bits are computed as the same, reducing the number of patterns. Then, a new method included the calculus of local variance to add contrast information to $LBP_{riu}$ \cite{guo2010rotation}. The measure is used to input weight for each pattern and finally feature vector LBPV is computed as:
\begin{equation}
\begin{split}
LBPV_{P,r} (k) = \sum_{i = 1}^{M} \sum_{j = 1}^{N} w(LBP_{P,r}^{riu} (i,j, k)) k \in [0,K]
\\ 
\\
w(LBP_{P,r} (i,j), k) = \left\{\begin{matrix} 
VAR_{P,r}(i,j), LBP_{P,r}^{riu} (i,j) = k\\ 0, \text{ otherwise} 
\end{matrix}\right.
\end{split}
\end{equation}
where K is the maximum grayscale level of the image, M and N are the number of rows and columns respectively and parameters P and r are set as the same of LBP, 8 and 1 respectively.

\item \textbf{Gray Level Difference Matrix} (GLDM):

GLDM first calculates a difference image, $f_d$, according to the equation: 

\begin{equation}
f_d(x,y) = f(x,y) - f(x + \Delta x, y + \Delta y),\end{equation} 

where f is the input image x, y are positions in image f with 1 $\leq$ x $\leq$ M and 1 $\leq$ y $\leq$ N (M and N are image dimensions). Then, a feature vector is computed by the concatenation of contrast, angular second moment, entropy and mean calculated from the probability function of $f_d$ \cite{weszka1976comparative}. Also, $\Delta x $ and $\Delta y$, parameters of the method, are set as \{(1,1),(2,2), (5,5)\} for the experiments. 

\item \textbf{Gray Level Co-occurrence  Matrix} (GLCM):

This statistical method computes the frequency between neighboring pixels \cite{haralick1979statistical}. In a co-occurrence matrix each position (p,q) represents the frequency that two pixels p and q separated by a distance $d$ and angle $\theta$ appears in the original image. These two arguments are parameters of the method are converted to displacements in the image as $\Delta x$ and $\Delta y$ used as $\Delta x$ =  $\Delta y$ = \{0,-1,1,2,-2\} in this paper. Then, correlation, energy, contrast and homogeneity are computed from co-occurrence matrix and concatenated to represent the image.

\item \textbf{Fourier Descriptor}:

Fourier transform changes the domain of the image f from spatial to frequency domain (represented by F) \cite{zahn1972fourier}. Therefore, features are taken from analyses of the spectrum and there are two main types of attributes that can be extracted: energy from circular rings Ea and energy from circular sections Eb \cite{azencott1997texture}:

\begin{equation}
E_{a_{ij}} = \int_{0}^{\lambda_i} \int_{\theta_j}^{\theta_{j+1}} |F(\lambda, \theta)|^2 d\theta d\lambda \label{eq:Ea}
\end{equation}

\begin{equation}
E_{b_{i}} = \int_{0}^{\lambda_i} \int_{0}^{2 \pi} |F(\lambda, \theta)|^2 d\theta d\lambda  \label{eq:Eb}
\end{equation}

where $\lambda$ and $\theta$ are polar coordinates of frequency domain. Also, 1 $\leq$ i $\leq$ r and 1 $\leq$ j $\leq$ N with r and N related to radius and orientations to delimit regions and used as 4 and 7 respectively in this paper. Energies of each region are used as characteristics of the image.

\item \textbf{Gabor Descriptor}:

When Fourier is used, spatial information is lost. Then, Gabor filters are a way to overcome this problem due to its combination of spatial and spectrum analyses \cite{daugman1988complete}. It first creates a bank of filters to represent a variety of features which are convolved with the image. Energy function is calculated from responses of each region in the image to these patterns and used as features. In this paper, filters are created according to 8 scales and 5 orientations.

\end{itemize}
\section{Proposed Method} \label{sec:proposal}

The goal of this work is to enhance characteristics of the original dataset enabling feature extraction methods to work better finding peculiarities of the scene to create robust representations. As said before, there are many ways to transform an image to improve its features. However, in this paper, binary distance transform is applied on original dataset to create a set of new enhanced images which can be used later to improve representation. 

Images from texture datasets are in grayscale. Thus, a binarization must be performed to transform them from grayscale to logical images. To reach this, for each iteration $i$ a binarization is done on image I of the dataset according to the following equation: 

\begin{equation}
I_i(x,y) = \left\{\begin{matrix}
1, \text{ if } I(x,y) \leq i \\0, \text{otherwise}
\end{matrix}\right.\label{eq:binarization}
\end{equation}

Then, for each logical image $I_i$ binary EDT is applied: 

\begin{equation}
I_{i}^{edt} = edt(I_i) \label{eq:edt}
\end{equation}

Finally, both distance image $I_{i}^{edt}$, output from iteration $i$ and original images are converted to feature vectors $v_d$ and $v_{d}^{i}$ respectively by descriptor $d$.  These vectors jointed are used to classify images. Therefore, for each experiment $n_i$ recognition rates are obtained using instead of only the original image, the addition of information from distance image $I_{i}^{edt}$. Our approach analyses which one of these iterations input most relevant characteristics to improve feature extraction. 

In this work, six different descriptors described in Section \ref{sec:features} are used along two classifiers to obtain a quantitative measure for the approach: KNN (k = 1) and Naive Bayes. Moreover, in experiments 150 binarizations are performed ($n_i$ = 150 and 1 $\leq$ i $\leq$ 150) meaning that for each combination, one descriptor and one classifier, 150 texture recognition are achieved

\begin{figure*}%
  \centering
  \includegraphics[scale=0.8]{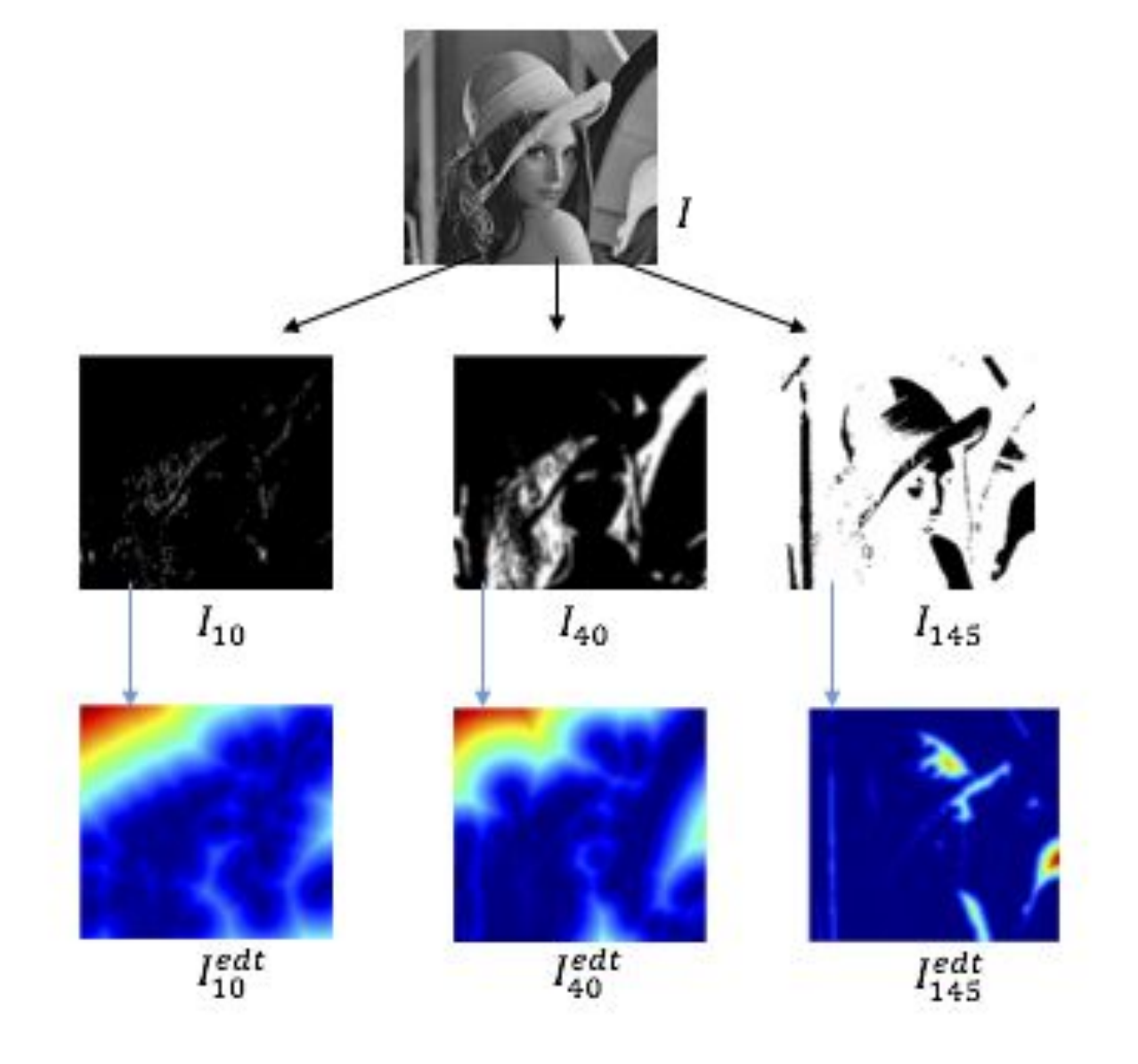} 
    \caption{From original image at first row, a set of binarizations are performed. Then, distance transform is applied as showed in third line (colors were added on transformed images to improve visualization)}  \label{fig:evolution}
\end{figure*}

\begin{figure*}%
  \centering
  \includegraphics[scale=0.6]{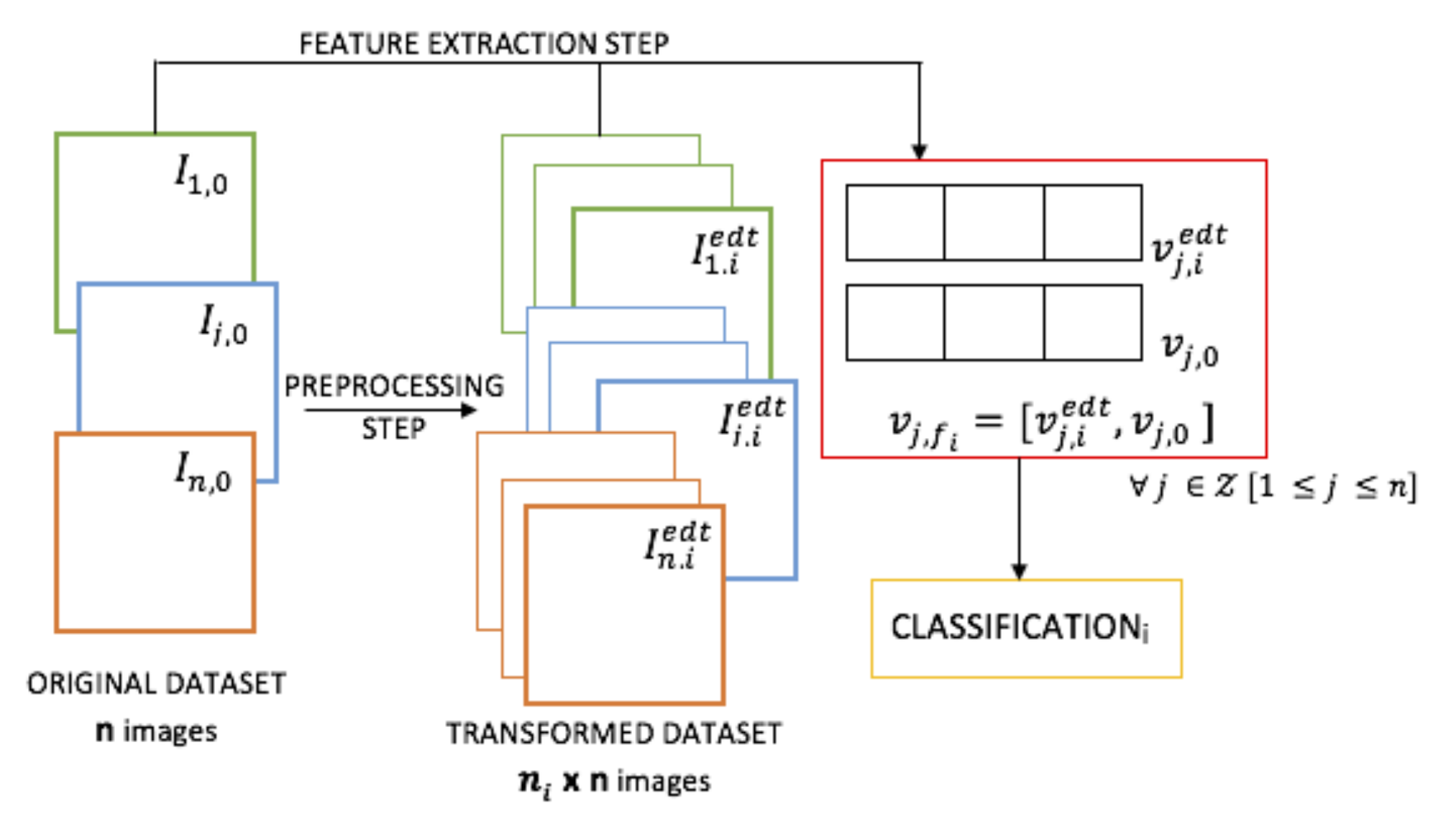} 
    \caption{Diagram of the proposed method. Three steps are performed, first images suffer a set of binarizations and resulted images are transform by EDT. Then all images from iteration $i$ and original ones are converted to a unidimensional representation by descriptor $d$ and finally classification is executed.}  \label{fig:diagram}
\end{figure*}

Figure \ref{fig:evolution} shows the binarization step and the resulting application of distance transform on these images. Also, Figure \ref{fig:diagram} shows a diagram of the proposed method.

\section{Results}\label{sec:results}

With the amount of results obtained (150 for each combination) it is impractical show all rates in this paper. Therefore, in result tables only the best iteration $i$ is showed. Also, to test the approach four dataset are used: Outex (1360 images in 68 classes) \cite{ojala2002outex}, Usptex (2292 images in 191 classes) \cite{backes2012color}, Vistex (864 images in 54 classes) \cite{pickard1995vistex} and Brodatz (1110 images in 111 classes) \cite{brodatz1966textures}. First three datasets contain images of size 128x128 with a variety of textures including differences in scales, illumination and view point. For these datasets, original images are colored but converted to grayscale to the experiments. Brodatz, otherwise, is a grayscale dataset with size 200x200 and no variations in scale, illumination and view point. 

These datasets are very common for texture recognition and each one of them contains their own characteristics that somehow makes texture classification complex. Therefore, results table (Tables \ref{table:outex}, \ref{table:brodatz},\ref{table:vistex}, \ref{table:usptex}) are separated by dataset and we can see which combination works better for each set of images. 

Overall, our proposed showed higher results in comparison to traditional approach. Although some results are lower compared to descriptors applied in the original dataset, it only happens for some combinations showing that binary EDT should not be used as a preprocessing method for GLDM, GLCM and LBPV. The last two combinations can be performed when Naive Bayes is used where gains in texture classification are noticed with our approach. 

Also, it is important to emphasize that to consider if the combination is bad, it is necessary to analyze the results according to each set. For Outex (Table \ref{table:outex}), for instance, the addition of one transformed image improved all feature descriptors for both classification methods. The results for Gabor filters was increased in up to 10.15\% with this new approach. Furthermore, the best recognition rate for Outex was reached when the distance image of iteration 67 was combined to improve representation of LBP. While traditional approach  had a success rate of 80.81\%, the proposed obtained 83.24\% considering same parameters and classifier.

\begin{table}[h!]
\centering
\caption{Results obtained testing the approach in Outex dataset. Table compares results from proposed approach and traditional texture classification by different descriptors. Two classifiers are tested: KNN (k = 1) and Naive Bayes. Also, cross validation (k-fold, k = 10) is to evaluated to understand the capacity of generalization of the method}
\label{table:outex}
\resizebox{\columnwidth}{!}{%
\begin{tabular}{|l|ll|ll|}
\hline
\multicolumn{5}{|c|}{Outex Dataset}                                                  \\ \hline
\multirow{2}{*}{} & \multicolumn{2}{l|}{KNN, k = 1} & \multicolumn{2}{l|}{Naive Bayes} \\ \cline{2-5} 
                  &            & best $i$           &             & best $i$           \\ \hline
LBP               &     72.50 (2.48)& -                 &      80.81 (3.47)& -                \\
EDT + LBP      &          \textbf{79.63} (2.58)&57                 &      \textbf{83.24 }(3.78)&67                 \\ \hline
LBPV              &     75.59 (4.20)& -           &    59.26 (4.13)& -                \\
EDT + LBPV    & \textbf{78.46} (3.80)&51  &\textbf{69.56} (3.75)&58                \\ \hline
GLCM              &    \textbf{72.72} (5.19)& -     &     62.35 (4.63)& -                \\
EDT+ GLCM     & \textbf{72.72} (5.19)&1
&      \textbf{73.09} (2.85)&58               \\ \hline
GLDM     &    74.04 (3.72)& - &     59.19 (4.72)& -    \\
EDT+ GLDM  & \textbf{79.85} (2.10)&57   &  \textbf{72.72} (3.63)&6 \\ \hline
Fourier           &    68.75 (2.76)& -             &       56.62 (3.70)& -       \\
EDT + Fourier  &     \textbf{72.65} (3.29)&69 &            \textbf{65.74} (2.92)&56     \\ \hline
Gabor             &      72.06 (2.11)& -             &   65.22 (2.48)& -               \\
EDT + Gabor    &    \textbf{78.09} (3.75)&51                 &  \textbf{75.37} (3.40)&56               \\ \hline
\end{tabular}%
}
\end{table}

For Brodatz (Table \ref{table:brodatz}), the simplest dataset, the best result was also obtained by a combination of our approach with LBP reaching a new success rate of 96.58\%. In this case, only the combination with EDT plus GLCM and EDT plus GLDM (by KNN) were lower compared to classical approach. GLDM was improved in 6.59\% and GLCM in 4.24\% when Naive Bayes is used instead of the lazy classifier. 

\begin{table}[h!]
\centering
\caption{Results obtained testing the approach in Brodatz dataset. Table compares results from proposed approach and traditional texture classification by different descriptors. Two classifiers are tested: KNN (k = 1) and Naive Bayes. Also, cross validation (k-fold, k = 10) is to evaluated to understand the capacity of generalization of the method}
\label{table:brodatz}
\resizebox{\columnwidth}{!}{%
\begin{tabular}{|l|ll|ll|}
\hline
\multicolumn{5}{|c|}{Brodatz Dataset}                                                  \\ \hline
\multirow{2}{*}{} & \multicolumn{2}{l|}{KNN, k = 1} & \multicolumn{2}{l|}{Naive Bayes} \\ \cline{2-5} 
                  &            & best $i$           &             & best $i$           \\ \hline
LBP&93.96(1.70)&  -                   &           95.05(1.60)& -                  \\
EDT + LBP&\textbf{96.49} (0.85)&91 
&   \textbf{96.58} (1.59)&132   \\ \hline
LBPV&  \textbf{94.23}(1.76) &   - 
& 91.26(2.72)& -            \\
EDT + LBPV&   93.42 (1.80)     &    126 &  \textbf{91.62} (2.46)&89     \\ \hline
GLCM&  \textbf{90.00}(2.77)  &   - &  84.77(2.80)& -  \\
EDT + GLCM&   88.65 (2.72)&76  & \textbf{89.01} (3.15)&70  \\ \hline
GLDM& \textbf{90.36}(2.04)& -&
    79.10(3.50)  & - \\
EDT + GLDM&87.57 (3.53)&77  &
\textbf{85.86} (4.74)   &   89 \\ \hline
Fourier  & 85.32(2.40)& -    & 81.44(2.95)& -     \\
EDT + Fourier& \textbf{86.67 }(1.94)&136  &  \textbf{83.33} (2.29)&128  \\ \hline
Gabor  & 90.90(2.10)& -   &    89.10(3.02)& - \\
EDT + Gabor& \textbf{92.43} (2.97)&148 &   \textbf{91.89 }(2.79)&89 \\ \hline
\end{tabular}%
}
\end{table}

The smallest dataset, Vistex (Table \ref{table:vistex}), also had the best result with the combination of the preprocessing method plus LBP, a 97.93\% (KNN) success rate. It can be perceived that it is best to use these combinations with Naive Bayes. In three combinations classified by KNN, the texture recognition is not improved (with LBPV, GLDM and GLCM), but for the same methods, classification is enhanced when Naive Bayes is used.

\begin{table}[h!]
\centering
\caption{Results obtained testing the approach in Vistex dataset. Table compares results from proposed approach and traditional texture classification by different descriptors. Two classifiers are tested: KNN (k = 1) and Naive Bayes. Also, cross validation (k-fold, k = 10) is to evaluated to understand the capacity of generalization of the method}
\label{table:vistex}
\resizebox{\columnwidth}{!}{%
\begin{tabular}{|l|ll|ll|}
\hline
\multicolumn{5}{|c|}{Vistex Dataset}                                                  \\ \hline
\multirow{2}{*}{} & \multicolumn{2}{l|}{KNN, k = 1} & \multicolumn{2}{l|}{Naive Bayes} \\ \cline{2-5} 
                  &            & best $i$           &             & best $i$           \\ \hline
LBP&    94.21 (2.74)        & -    &  95.49 (3.27)& -                     \\
EDT + LBP&      \textbf{97.92} (2.20)      &    92     &  \textbf{97.34} (2.29)&70                   \\ \hline
LBPV&    \textbf{92.82} (3.14)& -                &    79.75 (4.33)& -                 \\
EDT + LBPV&   91.78 (4.74)&67               &    \textbf{85.07} (3.70)&125                \\ \hline
GLCM&   \textbf{87.38} (3.56)& -                  &   73.03 (5.14)& -   \\
EDT + GLCM&    86.11 (5.41)&124                  & \textbf{82.41} (3.36)&72                \\ \hline
GLDM&    \textbf{88.89} (2.84)& -                  &             \textbf{77.20} (4.46)&53                  \\
EDT + GLDM&    87.04 (5.61)&54                 &            66.67 (5.64)& -                   \\ \hline
Fourier&     80.44 (3.60)& -                  &    71.99 (3.09)& -                  \\
EDT + Fourier&     \textbf{83.22} (2.77)&122         &  \textbf{79.40} (4.35)&68         \\ \hline
Gabor&   88.43 (1.93)& -                  &     84.14 (3.40)& -                 \\
EDT + Gabor&     \textbf{91.55} (2.60)&132                  &      \textbf{88.19} (2.69)&90               \\ \hline
\end{tabular}%
}
\end{table}

\begin{figure}%
  \centering
  \subfloat[LBP]{{\includegraphics[width=4cm]{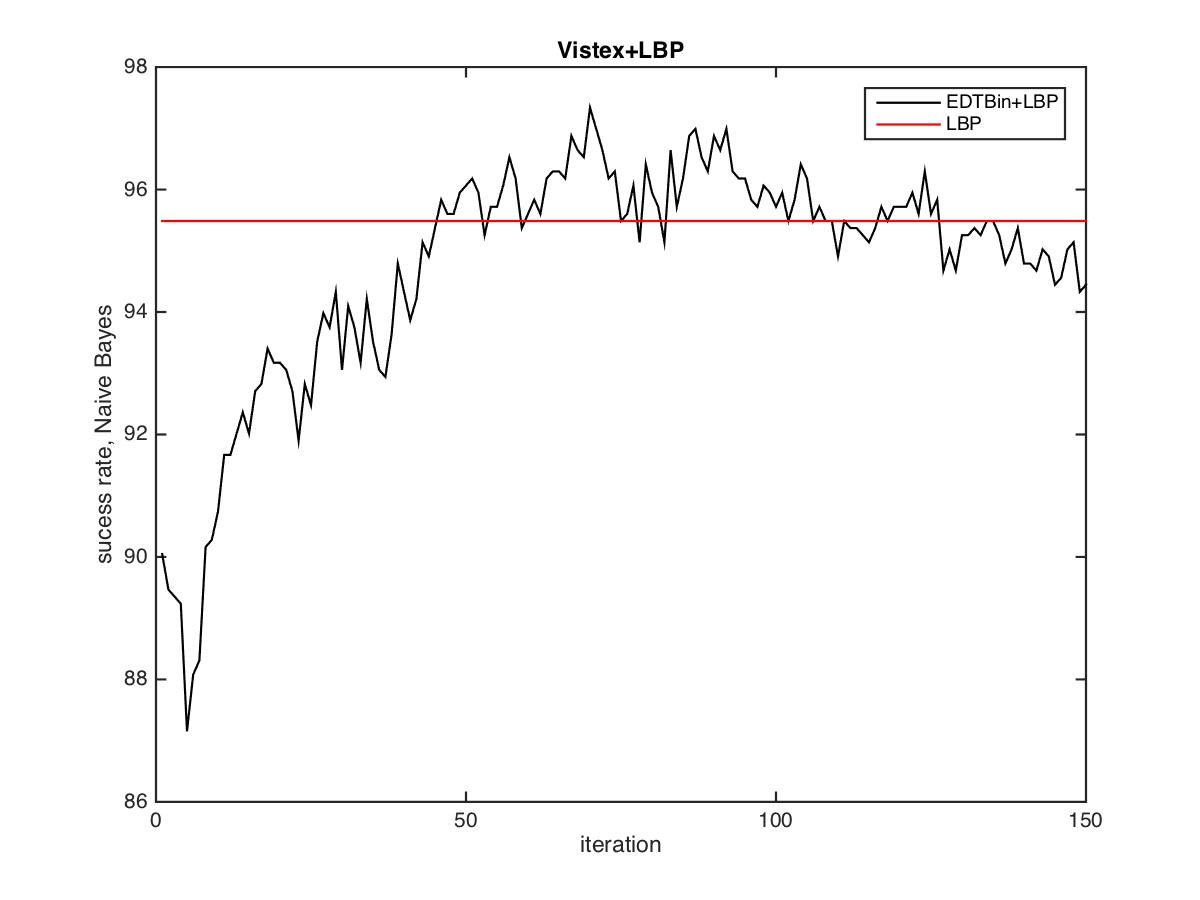} }}%
 \subfloat[LBPV]{{\includegraphics[width=4cm]{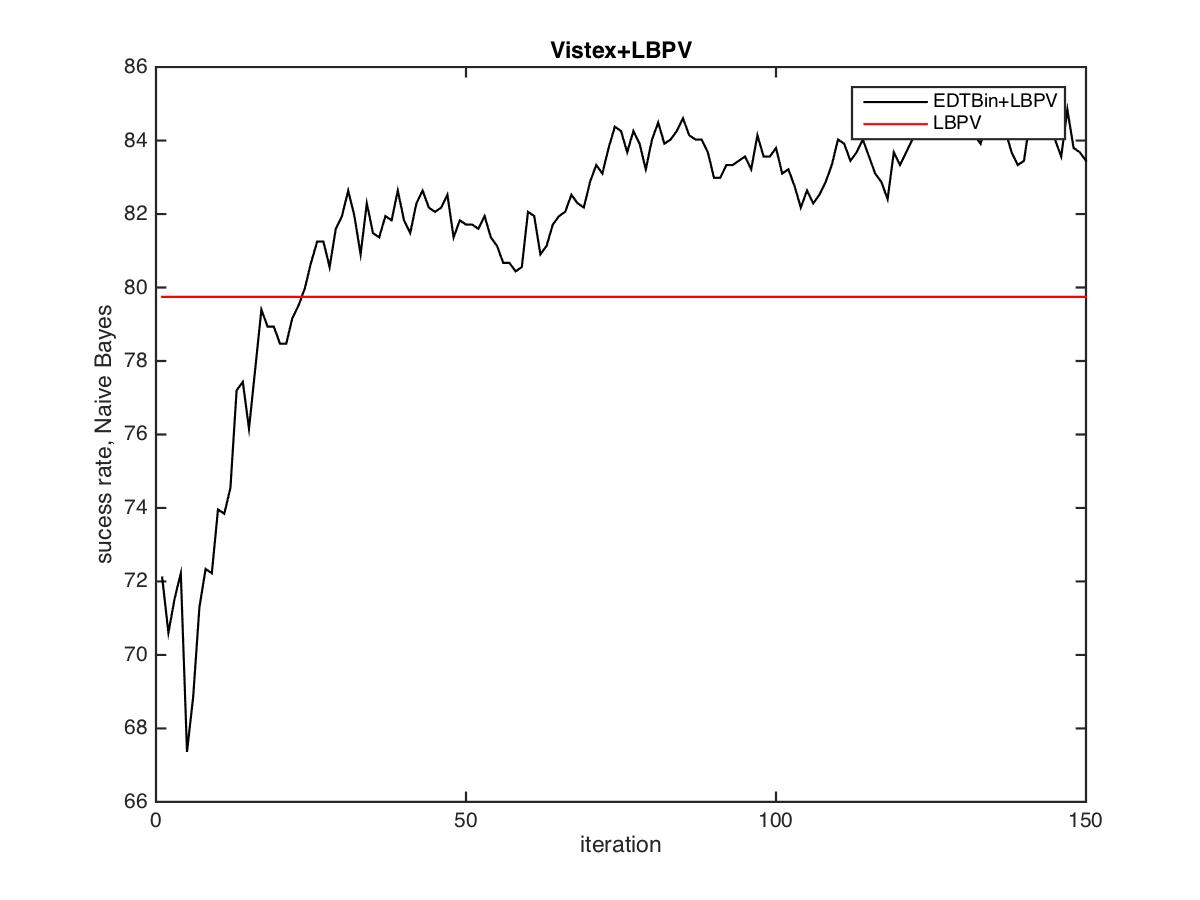} }}%
 \\
  \subfloat[GLCM]{{\includegraphics[width=4cm]{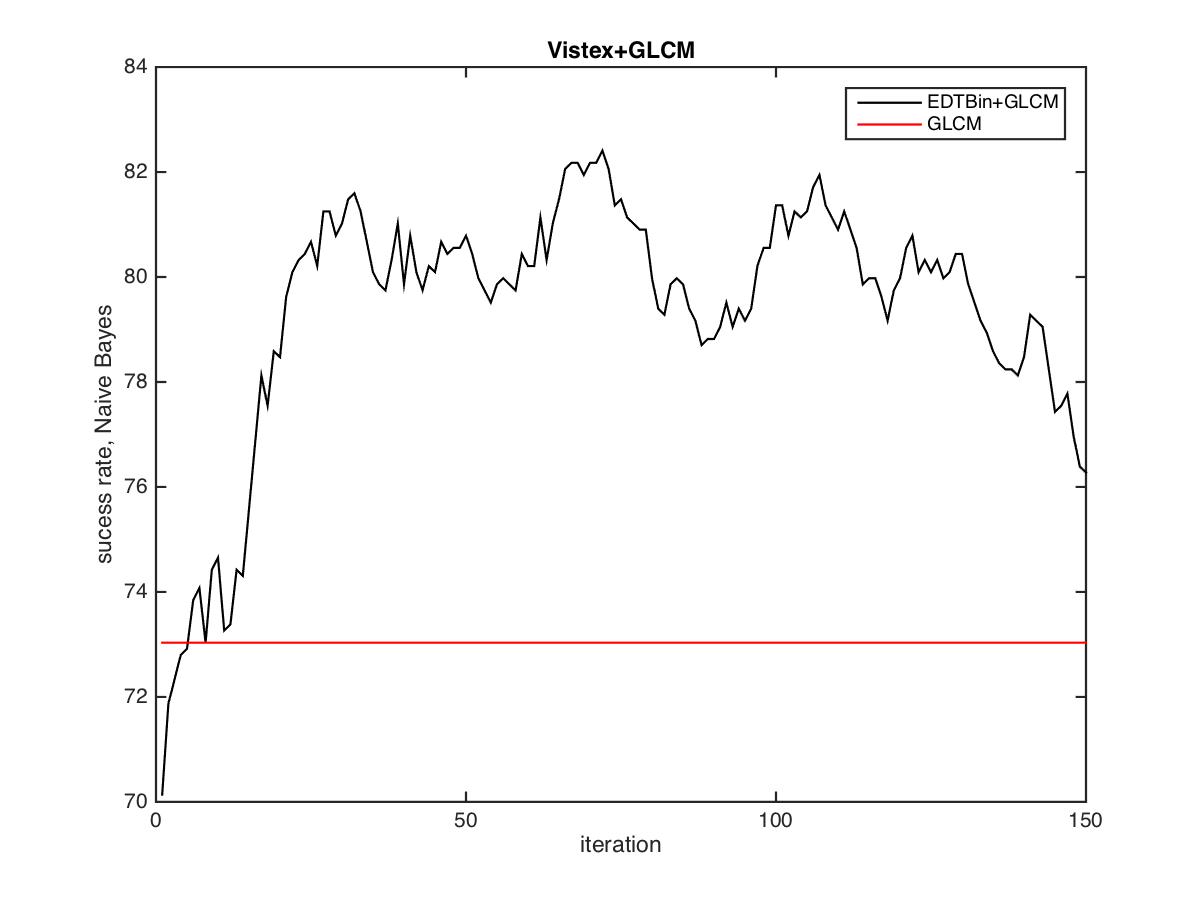} }}%
  \subfloat[GLDM]{{\includegraphics[width=4cm]{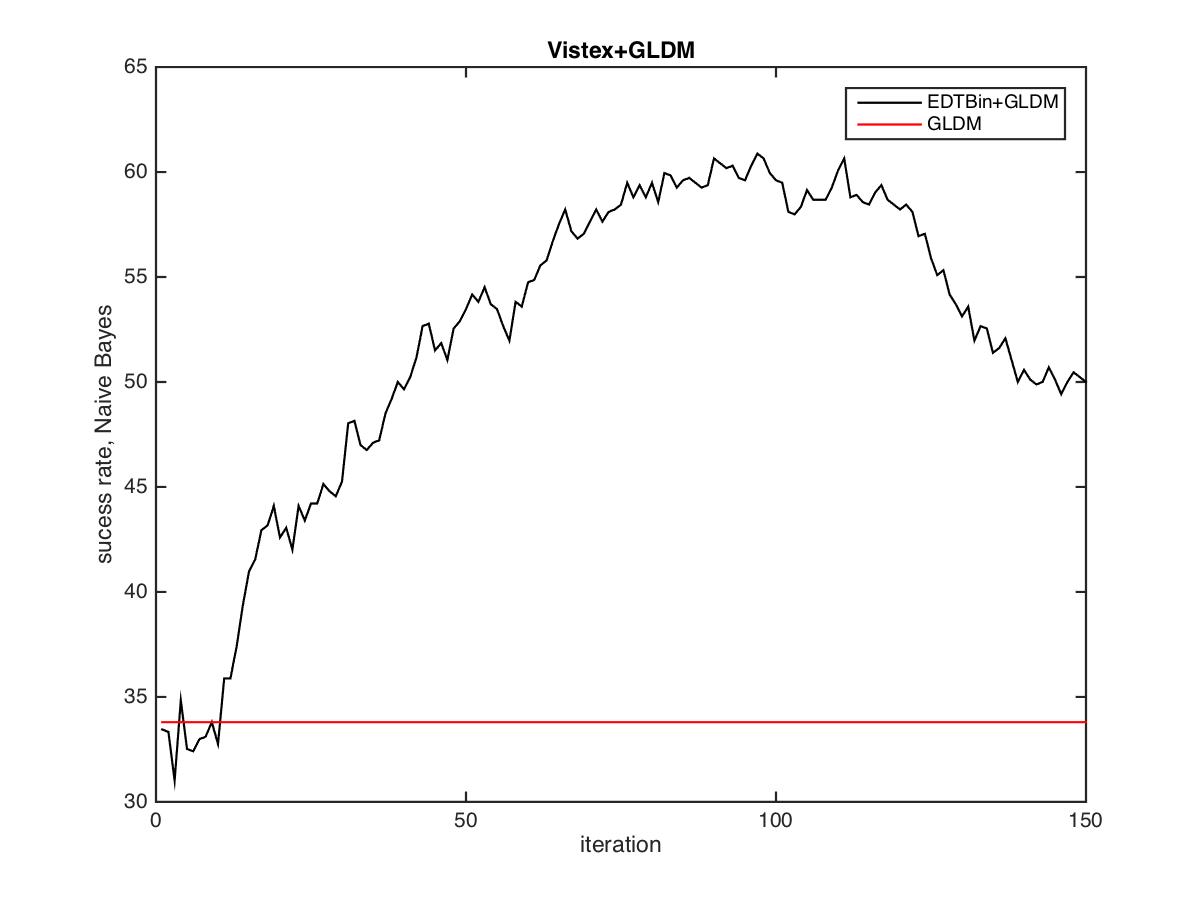} }}%
  \\
  \subfloat[Fourier]{{\includegraphics[width=4cm]{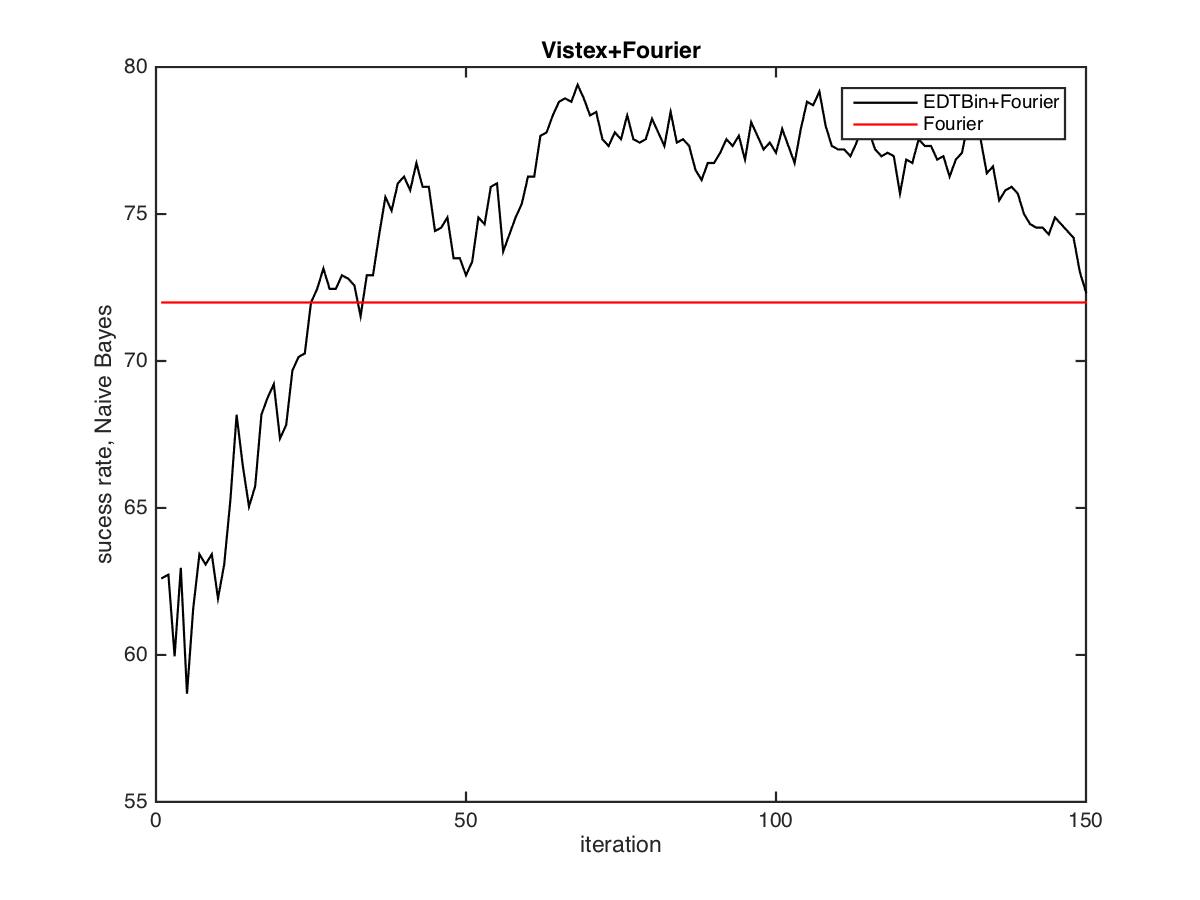} }}%
  \subfloat[Gabor]{{\includegraphics[width=4cm]{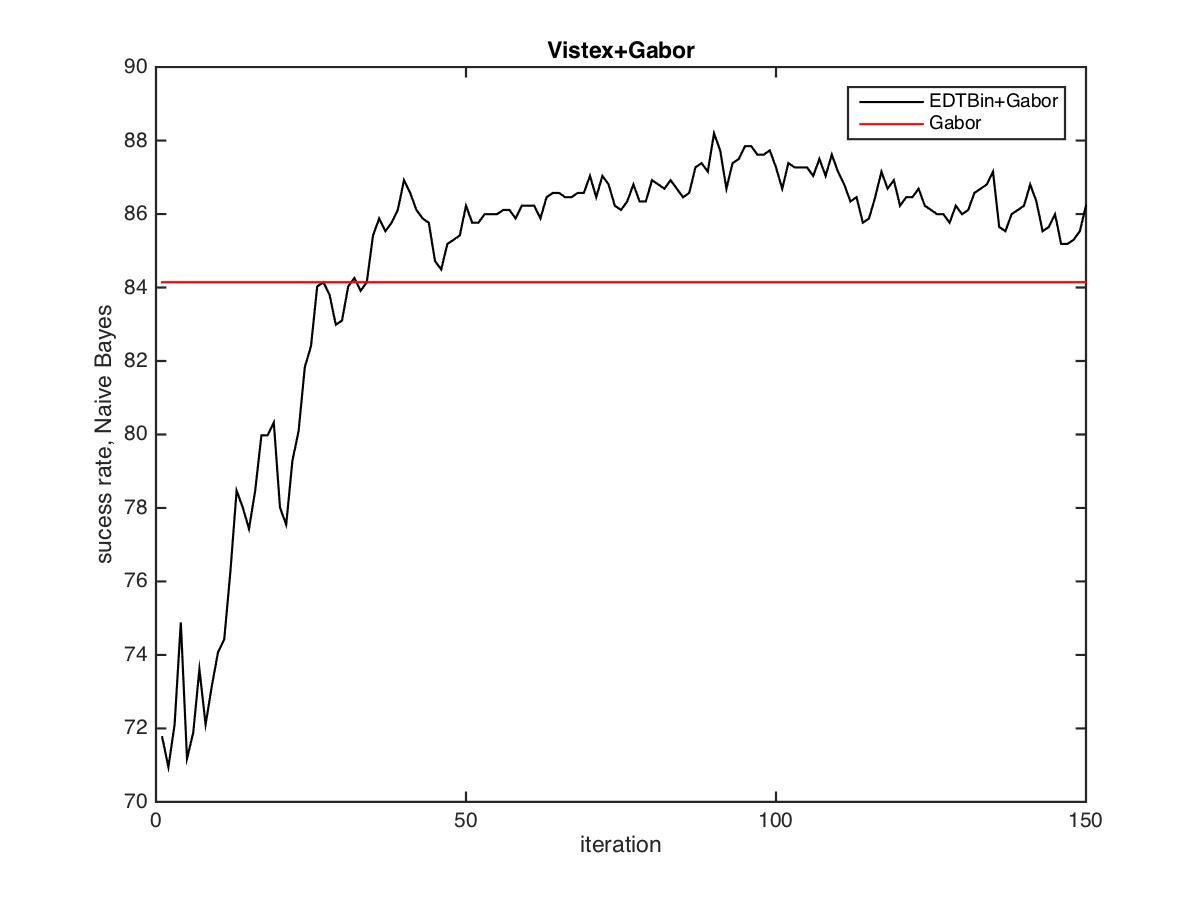} }}%
    \caption{Results of each iteration when preprocessing methods are combined with descriptors for Vistex dataset (Naive Bayes)} \label{plot:vistex}
\end{figure}

For this dataset, Figure \ref{plot:vistex} shows the results for all iterations $i$ and relative success rates by Naive Bayes. From all plots it is possible to see that GLCM and GLDM are the most improved methods for this set. However, all methods are overall enhanced by this approach, especially in higher iterations. 

Finally, the last and biggest dataset, Usptex was increased in up to 7.02\%. Same analyses with which classifier is better to use in these combinations: Naive Bayes. Descriptors suffers a higher improvement when this classifier is associated. In addition, highest recognition rate is hit with LBP considering the use of iteration 134 to add robustness to representation. Results with KNN (k = 1) is increased in 6.93\% going from 73.65\% to 80.58\%. 

\begin{table}[]
\centering
\caption{Results obtained testing the approach in Usptex dataset. Table compares results from proposed approach and traditional texture classification by different descriptors. Two classifiers are tested: KNN (k = 1) and Naive Bayes. Also, cross validation (k-fold, k = 10) is to evaluated to understand the capacity of generalization of the method}
\label{table:usptex}
\resizebox{\columnwidth}{!}{%
\begin{tabular}{|l|ll|ll|}
\hline
\multicolumn{5}{|c|}{Usptex Dataset}                                                  \\ \hline
\multirow{2}{*}{} & \multicolumn{2}{l|}{KNN, k = 1} & \multicolumn{2}{l|}{Naive Bayes} \\ \cline{2-5} 
                  &            & best $i$           &             & best $i$           \\ \hline
LBP&   73.65 (2.47)         &        -    &   78.71 (1.44)& -                    \\
EDT + LBP&  \textbf{80.58} (2.09)   &  134     &    \textbf{80.50} (2.93)&126                \\ \hline
LBPV&  \textbf{73.30} (3.62)& -         &     60.60 (1.76)& -                \\
EDT + LBPV&   73.08 (3.57)&1                &    \textbf{63.44} (3.37)&139                 \\ \hline
GLCM&  \textbf{63.74} (4.23)& -   &  47.86 (3.23)& -             \\
EDT + GLCM&56.54 (3.24)&136           &    \textbf{55.72} (3.02)&144\\\hline
GLDM&   71.86 (1.90)& -      &     52.71 (2.57)& -                     \\
EDT + GLDM&    \textbf{72.16} (2.04)&1                  &             \textbf{61.43} (3.51)&135                  \\ \hline
Fourier&56.85 (3.23)& - & 48.43 (2.98)& -                \\
EDT + Fourier&      \textbf{57.72} (2.73)&136  &     \textbf{53.88} (2.79)&138    \\ \hline
Gabor&       70.59 (2.71)& -       & 62.52 (2.85)& -                 \\
EDT + Gabor&      \textbf{70.90} (2.83)&1                 &    \textbf{65.84} (2.83)&144\                  \\ \hline
\end{tabular}%
}
\end{table}

Furthermore, it was observed that the iteration that added most representation for classification is very particular of each combination. Therefore, it is important to perform the preprocessing step to analyze this aspect. Since the preprocessing method is very fast (methods was implemented in Matlab) and gains are considerable, the increasing of cost in this proposal compared to traditional approach is minimal.

\section{Conclusion} \label{sec:conclusions}

This paper proposed a combination of the original dataset with transformed images by binary euclidean distance transform. The method of preprocessing has two main steps: first, it thresholds the image according to a value $i$ and then distance transform is applied on these binary images. The main goal was check if the addition of a modified image increased feature extraction power of tested methods and to evaluate it, texture recognition was performed.

Also, four datasets were used and results shows that in general the addition of an image does improves texture recognition. For instance, Outex had an increase of 10.74\% of classification in comparison with feature extraction by LBP applied only in the original dataset. 

Among all feature extraction methods, the most benefited by the new proposal was GLCM (classified by Naive Bayes) as can be noticed in Figure \ref{plot:vistex} and tables of results. However, the method that reached the highest result for all datasets is the combination of binary distance transform and LBP. 

Due to differences in dataset and the way that descriptors analyses images, this approach must be observed according to each dataset. However, as can be seen in results, most of the combinations improves all datasets. From this results, new descriptors and different methods can be tested to show the power of the addition of preprocessed images to empower feature extraction and consequently increase texture recognition. 

\section*{Acknowledgments}

The authors acknowledges support from CNPq (National Counsel of Technological and Scientific Development) (Grant Nos. 132409/2014-3, 307797/2014-7 and 484312/2013-8) and FAPESP (Grant No. 14/08026-1).




\begin{thebibliography}{10}

\bibitem{tuceryan1993texture}
Mihran Tuceryan, Anil~K Jain, et~al.
\newblock Texture analysis.
\newblock {\em Handbook of pattern recognition and computer vision},
  2:207--248, 1993.

\bibitem{kim1999statistical}
Jong~Kook Kim and Hyun~Wook Park.
\newblock Statistical textural features for detection of microcalcifications in
  digitized mammograms.
\newblock {\em Medical Imaging, IEEE Transactions on}, 18(3):231--238, 1999.

\bibitem{haralick1979statistical}
Robert~M Haralick.
\newblock Statistical and structural approaches to texture.
\newblock {\em Proceedings of the IEEE}, 67(5):786--804, 1979.

\bibitem{garcia2014local}
Oscar Garc{\'\i}a-Olalla, Enrique Alegre, Laura Fern{\'a}ndez-Robles, and
  V{\'\i}ctor Gonz{\'a}lez-Castro.
\newblock Local oriented statistics information booster (losib) for texture
  classification.
\newblock In {\em 2014 22nd International Conference on Pattern Recognition
  (ICPR)}, pages 1114--1119. IEEE, 2014.

\bibitem{costa2012efficient}
Alceu~Ferraz Costa, Gabriel Humpire-Mamani, and Agma Juci~Machado Traina.
\newblock An efficient algorithm for fractal analysis of textures.
\newblock In {\em Graphics, Patterns and Images (SIBGRAPI), 2012 25th SIBGRAPI
  Conference on}, pages 39--46. IEEE, 2012.

\bibitem{guo2010completed}
Zhenhua Guo, Lei Zhang, and David Zhang.
\newblock A completed modeling of local binary pattern operator for texture
  classification.
\newblock {\em Image Processing, IEEE Transactions on}, 19(6):1657--1663, 2010.

\bibitem{guo2010rotation}
Zhenhua Guo, Lei Zhang, and David Zhang.
\newblock Rotation invariant texture classification using lbp variance (lbpv)
  with global matching.
\newblock {\em Pattern recognition}, 43(3):706--719, 2010.

\bibitem{ojala2002multiresolution}
Timo Ojala, Matti Pietik{\"a}inen, and Topi M{\"a}enp{\"a}{\"a}.
\newblock Multiresolution gray-scale and rotation invariant texture
  classification with local binary patterns.
\newblock {\em Pattern Analysis and Machine Intelligence, IEEE Transactions
  on}, 24(7):971--987, 2002.

\bibitem{journaux2008texture}
Ludovic Journaux, Marie-France Destain, Johel Miteran, Alexis Piron, and
  Frederic Cointault.
\newblock Texture classification with generalized fourier descriptors in
  dimensionality reduction context: An overview exploration.
\newblock In {\em Artificial Neural Networks in Pattern Recognition}, pages
  280--291. Springer, 2008.

\bibitem{daugman1988complete}
John~G Daugman.
\newblock Complete discrete 2-d gabor transforms by neural networks for image
  analysis and compression.
\newblock {\em Acoustics, Speech and Signal Processing, IEEE Transactions on},
  36(7):1169--1179, 1988.

\bibitem{florindo2016local}
Joao~B Florindo and Odemir~M Bruno.
\newblock Local fractal dimension and binary patterns in texture recognition.
\newblock {\em Pattern Recognition Letters}, 78:22--27, 2016.

\bibitem{da2015feature}
Marcos~William da~Silva~Oliveira, N{\'u}bia~Rosa da~Silva, Antoine Manzanera,
  and Odemir~Martinez Bruno.
\newblock Feature extraction on local jet space for texture classification.
\newblock {\em Physica A: Statistical Mechanics and its Applications},
  439:160--170, 2015.

\bibitem{materka1998texture}
Andrzej Materka, Michal Strzelecki, et~al.
\newblock Texture analysis methods--a review.
\newblock {\em Technical university of lodz, institute of electronics, COST B11
  report, Brussels}, pages 9--11, 1998.

\bibitem{gonccalves2016texture}
Wesley~Nunes Gon{\c{c}}alves, N{\'u}bia~Rosa da~Silva, Luciano
  da~Fontoura~Costa, and Odemir~Martinez Bruno.
\newblock Texture recognition based on diffusion in networks.
\newblock {\em Information Sciences}, 364:51--71, 2016.

\bibitem{da2015improved}
N{\'u}bia~Rosa da~Silva, Pieter Van~der Wee{\"e}n, Bernard De~Baets, and
  Odemir~Martinez Bruno.
\newblock Improved texture image classification through the use of a
  corrosion-inspired cellular automaton.
\newblock {\em Neurocomputing}, 149:1560--1572, 2015.

\bibitem{florindo2016texture}
Jo{\~a}o~Batista Florindo and Odemir~Martinez Bruno.
\newblock Texture analysis by fractal descriptors over the wavelet domain using
  a best basis decomposition.
\newblock {\em Physica A: Statistical Mechanics and its Applications},
  444:415--427, 2016.

\bibitem{florindo2012fractal}
Jo{\~a}o~Batista Florindo and Odemir~Martinez Bruno.
\newblock Fractal descriptors based on fourier spectrum applied to texture
  analysis.
\newblock {\em Physica A: Statistical Mechanics and its Applications},
  391(20):4909--4922, 2012.

\bibitem{florindo2012comparative}
Jo{\~a}o~Batista Florindo, Andr{\'e}~Ricardo Backes, M{\'a}rio de~Castro, and
  Odemir~Martinez Bruno.
\newblock A comparative study on multiscale fractal dimension descriptors.
\newblock {\em Pattern Recognition Letters}, 33(6):798--806, 2012.

\bibitem{florindo2013texture}
Jo{\~a}o~Batista Florindo and Odemir~Martinez Bruno.
\newblock Texture analysis by multi-resolution fractal descriptors.
\newblock {\em Expert Systems with Applications}, 40(10):4022--4028, 2013.

\bibitem{gonzalez2008digital}
Rafael~C Gonzalez and Richard~E Woods.
\newblock Digital image processing.
\newblock {\em Nueva Jersey}, 2008.

\bibitem{machado2013partial}
Bruno~Brandoli Machado, Dalcimar Casanova, Wesley~Nunes Gon{\c{c}}alves, and
  Odemir~Martinez Bruno.
\newblock Partial differential equations and fractal analysis to plant leaf
  identification.
\newblock In {\em Journal of Physics: Conference Series}, volume 410, page
  012066. IOP Publishing, 2013.

\bibitem{nanni2015improving}
Loris Nanni, Sheryl Brahnam, Stefano Ghidoni, and Emanuele Menegatti.
\newblock Improving the descriptors extracted from the co-occurrence matrix
  using preprocessing approaches.
\newblock {\em Expert Systems With Applications}, 42(22):8989--9000, 2015.

\bibitem{gadermayr2016making}
Michael Gadermayr and Andreas Uhl.
\newblock Making texture descriptors invariant to blur.
\newblock {\em EURASIP journal on image and video processing}, 2016(1):1--9,
  2016.

\bibitem{parker2010algorithms}
Jim~R Parker.
\newblock {\em Algorithms for image processing and computer vision}.
\newblock John Wiley \& Sons, 2010.

\bibitem{vincent1991watersheds}
Luc Vincent and Pierre Soille.
\newblock Watersheds in digital spaces: an efficient algorithm based on
  immersion simulations.
\newblock {\em IEEE transactions on pattern analysis and machine intelligence},
  13(6):583--598, 1991.

\bibitem{chin2001vision}
Yew~Tuck Chin, Han Wang, Leng~Phuan Tay, Hui Wang, and William~YC Soh.
\newblock Vision guided agv using distance transform.
\newblock In {\em Proceedings of the 32nd ISR (International Symposium on
  Robotics)}, volume~19, page~21. Citeseer, 2001.

\bibitem{rosenfeld1968distance}
Azriel Rosenfeld and John~L Pfaltz.
\newblock Distance functions on digital pictures.
\newblock {\em Pattern recognition}, 1(1):33--61, 1968.

\bibitem{maurer2003linear}
Calvin~R Maurer~Jr, Rensheng Qi, and Vijay Raghavan.
\newblock A linear time algorithm for computing exact euclidean distance
  transforms of binary images in arbitrary dimensions.
\newblock {\em Pattern Analysis and Machine Intelligence, IEEE Transactions
  on}, 25(2):265--270, 2003.

\bibitem{ojala2000gray}
Timo Ojala, Matti Pietik{\"a}inen, and Topi M{\"a}enp{\"a}{\"a}.
\newblock Gray scale and rotation invariant texture classification with local
  binary patterns.
\newblock In {\em European Conference on Computer Vision}, pages 404--420.
  Springer, 2000.

\bibitem{weszka1976comparative}
Joan~S Weszka, Charles~R Dyer, and Azriel Rosenfeld.
\newblock A comparative study of texture measures for terrain classification.
\newblock {\em IEEE Transactions on Systems, Man, and Cybernetics},
  4(SMC-6):269--285, 1976.

\bibitem{zahn1972fourier}
Charles~T Zahn and Ralph~Z Roskies.
\newblock Fourier descriptors for plane closed curves.
\newblock {\em IEEE Transactions on computers}, 100(3):269--281, 1972.

\bibitem{azencott1997texture}
Robert Azencott, Jia-Ping Wang, and Laurent Younes.
\newblock Texture classification using windowed fourier filters.
\newblock {\em IEEE Transactions on Pattern Analysis and Machine Intelligence},
  19(2):148--153, 1997.

\bibitem{ojala2002outex}
Timo Ojala, Topi Maenpaa, Matti Pietikainen, Jaakko Viertola, Juha Kyllonen,
  and Sami Huovinen.
\newblock Outex-new framework for empirical evaluation of texture analysis
  algorithms.
\newblock In {\em Pattern Recognition, 2002. Proceedings. 16th International
  Conference on}, volume~1, pages 701--706. IEEE, 2002.

\bibitem{backes2012color}
Andr{\'e}~Ricardo Backes, Dalcimar Casanova, and Odemir~Martinez Bruno.
\newblock Color texture analysis based on fractal descriptors.
\newblock {\em Pattern Recognition}, 45(5):1984--1992, 2012.

\bibitem{pickard1995vistex}
R~Pickard, C~Graszyk, S~Mann, J~Wachman, L~Pickard, and L~Campbell.
\newblock Vistex database.
\newblock {\em Media Lab., MIT, Cambridge, Massachusetts}, 1995.

\bibitem{brodatz1966textures}
Phil Brodatz.
\newblock {\em Textures: a photographic album for artists and designers},
  volume~66.
\newblock Dover New York, 1966.

\end{thebibliography}



\end{document}